\def\year{2021}\relax
\documentclass[preprint]{article} 
\usepackage{aaai21}  
\usepackage{times}  
\usepackage{helvet} 
\usepackage{courier}  
\usepackage[hyphens]{url}  
\usepackage{graphicx} 
\usepackage{natbib}  
\usepackage{caption} 
\usepackage[export]{adjustbox}
\usepackage{pdflscape}
\usepackage{enumitem}

\usepackage{booktabs}
\usepackage{siunitx}
\usepackage[table]{xcolor}
\usepackage[ruled,vlined]{algorithm2e}
 \usepackage{subcaption} 
\SetKwInOut{Parameter}{parameter}

\title{Latent-CF: A Simple Baseline for Reverse Counterfactual Explanations}
\author{
    Rachana Balasubramanian, Sam Sharpe, Brian Barr, C. Bayan Bruss\\}
\affiliations{
    Center for Machine Learning, Capital One\\

    11 W 19th St\\
    New York, New York 10011\\
    {rachana.balasubramanian,bayan.bruss,brian.barr,samuel.sharpe}@capitalone.com

}

\begin{document}

\maketitle

\begin{abstract}
In the environment of growing concern about data, machine learning, and their use in high-impact decision making, the ability to explain a model's prediction is of paramount importance. High quality explanations are the first step in assessing fairness.  Counterfactual explanations answer the question "What is the minimal change to a data sample to alter its classification?"  They provide actionable, comprehensible explanations for the individual who is subject to decisions made from the prediction. A growing number of methods exist for generating counterfactual explanations, but it is unclear which is the best to use and why. We propose a baseline method for generating counterfactuals using gradient descent in the latent space of an autoencoder (AE). This simple but strong method can be used to understand more complex approaches to counterfactual generation. To aid comparison of counterfactual methods, we propose metrics to concretely evaluate the quality of the counterfactual explanations.  We compare our simple method against other approaches that search for counterfactuals in feature space or use latent space reconstruction as regularization. We show that latent space counterfactual search strikes a balance between the speed of basic feature perturbation  methods and the sparseness and authenticity of counterfactuals generated by more complex feature space techniques.
\end{abstract}


\noindent In response to increasing concern about algorithmic decision making, some governments have passed laws such as the General Data Protection Regulation (GDPR) to provide a "right to be informed" about system functionality in automated decision making processes. Though not explicit in the GDPR, the law encourages those creating algorithmic decision making systems to build trust and increase transparency around these systems \citep{wachter2017counterfactual}. As such, when considering the application of artificial intelligence in industrial settings, it is important to consider the information and agency afforded to those affected by the decisions of the AI system. In particular, it may be important to provide the affected party with a means to either contest the current outcome or change their behavior to ensure a better outcome in the future. 

One way to provide this means is through counterfactual explanations. A counterfactual explanation is a local explanation. For a given input to a machine learning model and its corresponding prediction, the counterfactual provides an alternative input that would have resulted in a different prediction. There have been a number of techniques proposed in recent years offering different ways of generating these counterfactuals \citep{molnar2019}. Counterfactual explanations trace their lineage to philosophy and psychology and are deeply intertwined with analyses of causality \cite{lewis1973, pearl2011algorithmization}. It is important to understand this lineage as it shapes the current approaches.

Counterfactual generation is a process that can happen in one of two directions. In philosophy and causal inference counterfactuals are part of a forward process. The goal is to understand how changes to a cause alters its downstream effects. In psychology, researchers have studied the counterfactuals in human cognition noting its development as early as two years old \cite{epstude2008functional}. Functional theories of counterfactual thinking suggest a reverse process. When there is a mismatch between expectations and the present situation, people work backward along the causal path until a way opens up to the desired outcome. This new path then informs behavior change. This view fits closely with the framework of reverse causal inference in statistics \cite{NBERw19614}.

In the domain of model-based counterfactual explorations, the common approach is aligned to the traditional forward looking causal structure. Namely, counterfactual approaches explore alternative inputs that would lead to changes in the model's predictions. The objective of a counterfactual is then given $f(X)$, for a specific $x_0$ where $f(x_0) = y_0$ generate $x_{cf}$ where $f(x_{cf}) = y_{cf}$. What is clear from this formulation is that there may be many possible paths to generating $x_{cf}$. As a result, forward looking counterfactuals seek to place constraints on the possible changes both for computational efficiency but also interpretability \cite{wachter2017counterfactual}. 

In this paper we propose a method for generating counterfactual explanations aligned to reverse causal inference and the functional theory of counterfactual thinking. We refer to our approach as \textit{Latent-CF}. Given a differentiable classifier and a dataset that has been used to train the classifier, Latent-CF can generate counterfactuals. Training a separate autoencoder on the same training data is necessary for our approach. For any data point and desired class confidence of the target counterfactual class, we traverse the latent space of the autoencoder from the original data point's encoded representation until the desired class probability is reached and then use the decoder to generate the corresponding counterfactual. In practical terms, the approach is simple but provides a foundation for viewing counterfactual explanations from a more functional perspective compared to existing forward oriented counterfactuals. 

In order to evaluate the quality of counterfactuals we use a framework similar to that of \citet{looveren2019interpretable}. We propose that a counterfactual explanation should be:

\begin{enumerate}
    \item In distribution - the proposed features should not have a low probability
of occurring
    \item Sparse in the number of changes it makes in the features
    \item Computationally efficient
\end{enumerate}


We conduct the following experiments to evaluate the quality of the counterfactuals generated by Latent-CF. First, using the criteria above, we compare Latent-CF to approaches that generate counterfactual explanations utilizing gradient descent (GD) or other optimization methods in feature space with varying loss and clipping strategies to encourage the explanations to be in-sample and sparse.  Second, we conduct a visual analysis of the counterfactual explanations. Through these experiments, we are able to show that Latent-CF provides a strong baseline to compare current (and future) methods in counterfactual generation.

\section{Previous Work}
There is a large body of existing work on \textit{post-hoc} explainability, see \citet{molnar2019} for an overview.  Many methods treat the creation of explanations as a problem of credit assignment --- for each input sample, provide the relative importance of the input features to that sample's prediction.  Early methods had roots in computer vision, where a common approach is to calculate sensitivities and show a heat map depicting which pixels are responsible for an image's classification. These methods \citep[e.g.][]{shrikumar2017} rely on using a zero information baseline.  The need for a baseline or reference population can become problematic if applied to tabular or other data where the definition of a baseline may not be clear or even exist.

The use of local surrogate models, such as LIME~\cite{ribeiro2016} and its extensions,  can overcome the need for global baselines to be established.  However, the perturbation method used does not ensure that the local surrogate is built on in-sample data.

Game theoretic approaches \cite{strumbelj2010}, \cite{datta2016}, \cite{lundberg2017} view the credit assignment task as a coalitional game amongst features. These methods baseline a feature's contribution against the average model prediction.  Game theoretic approaches are challenged by the exponential number of coalitions that must be evaluated.

Other researchers have developed methods that express the commonality of features that must be present or missing.  \citet{dhurandhar2018} try to find perturbations in feature space utilizing an autoencoder reconstruction loss to keep the explanations in sample. Their method seeks to find contrastive explanations, which describe what minimal set of features or characteristics must be missing to explain why it does not belong to an alternate class. 

Counterfactual explanations avoid the pitfalls of previous local attribution methods (no need for baselines, no approximation to game theoretic constructs, no need for universality in features).  All they require is a sample in need of explanation and a search method to find the decision boundary.  \citet{lash2016} maintain  sparsity  in  their  \textit{inverse  classification} methods  by  partitioning  the  features  into mutable and immutable categories, and imposing budgetary constraints on allowable changes to the mutable features.  \citet{laugel2018} advocate a sampling approach with their growing spheres method.

\citet{wachter2017counterfactual} proposed the use of counterfactual explanations to help the individuals impacted by a model based decision understand why a particular decision was reached, to provide grounds to contest adverse decisions,  and to  understand  what would result in an alternative decision.  \citet{looveren2019interpretable} explore loss functions to search for perturbations in feature space and introduce \textit{prototype} loss which encourages the counterfactual to be close to the average representation in latent space of $K$ training samples from the target class.  In contrast to our method which searches for a single counterfactual instance, \citet{mothilal2020} suggest generating a set of diverse counterfactuals.  \citet{mcgrath2018} used counterfactuals to explain credit application predictions.  They introduced weights, based on either global feature importance or nearest neighbors in an attempt to make them more interpretable.

Some recent counterfactual generation techniques search directly in a latent space. \citet{Pawelczyk_2020} construct a model agnostic technique that samples observations uniformly in $l_p$ spheres around the original point's representation in latent space.  \citet{joshi2019realistic} utilize a very similar approach to ours using a variational autoencoder (VAE) and traverse the latent space with gradient descent to find counterfactuals. They include an additional loss term in the feature space to encourage sparse changes. We choose to use the most simple manifestation of latent space generated counterfactuals---a basic autoencoder and loss including only score of a decoded latent representation---to convincingly illustrate the benefits of searching in a smaller representative space. 

\section{Methods}

\subsection{Latent-CF}
\begin{figure*}[hbt!]
    \centering
    \includegraphics[width=0.7\linewidth,trim=30 75 0 85]{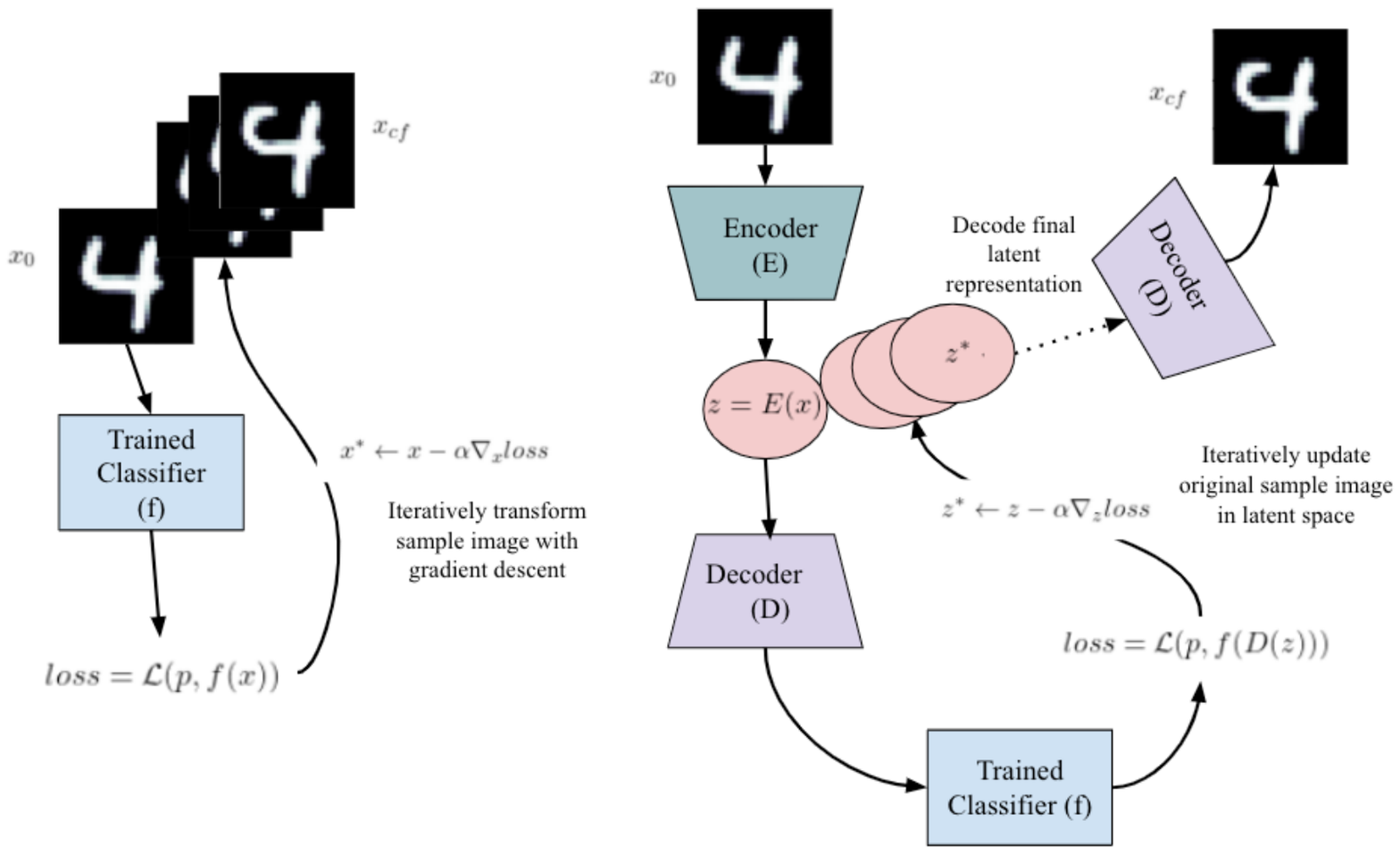}
    \caption{Comparing approaches for counterfactual generation between gradient descent in feature space (left) and Latent-CF gradient descent in the autoencoder latent space (right)}
    \label{fig:arch}
\end{figure*}
Feature space perturbation methods feed a sample $x_0$ through a classifier $f: X \rightarrow [0,1]$ and update it until $f(x)$ is close to $0.5$ (decision boundary) or some other target probability $p$ where the closeness to the boundary is measured by a loss function $\mathcal{L}$. 
Approaches in \citet{looveren2019interpretable} and \citet{dhurandhar2018} incorporate an autoencoder  in $\mathcal{L}$ to guide the search for counterfactual examples. However, Latent-CF, similar to \citet{joshi2019realistic}, directly searches near the latent representation of the encoder, $z = E(x)$, until the probability of the decoded sample, $f(D(z))$, is close to $p$. Our algorithm is detailed in Algorithm \ref{algo:b} and the architecture of Latent-CF is illustrated in Figure \ref{fig:arch}. We also perform experiments with a VAE in place of an autoencoder to examine what benefits, if any, a regularized latent space provides. 
\begin{algorithm}
\Parameter{$p$ probability of target counterfactual class (0.5 for decision boundary), $tol$ tolerance}
\KwIn{Instance to explain $x_0$, classifier $f$, encoder $E$, decoder $D$, $\mathcal{L}$ loss function}
\KwOut{$x_{cf}$ the counterfactual explanation}
Encode sample to latent space $z = E(x_0)$ \\
Calculate $loss = \mathcal{L}(p, f(D(z))$\\
\While{$loss > tol$}{
    $z \leftarrow z - \alpha \nabla_z loss$ \\
    Calculate $loss = \mathcal{L}(p, f(D(z))$
}
$x_{cf} = D(z)$
\caption{Latent-CF}
\label{algo:b}
\end{algorithm}

\subsection{Comparison Methods}
Our first comparison method, \textit{Feature GD} (FGD), uses gradient descent to directly perturb the feature space minimizing the $\ell_{2}$ distance to the decision boundary. Two other methods introduce some small changes to make iterative improvements over Feature GD. First, we add feature clipping after every gradient step in \textit{Feature GD + clip} (FGD+C) to ensure pixel values stay close to the training data domain.

We also implement \textit{Feature GD + MAD loss} (FGD+MAD), which encourages in-sample counterfactuals and sparse changes with Median Absolute Deviation (MAD) scaling loss, as developed by \citet{wachter2017counterfactual}. Instead of the $\ell_2$ loss, this method uses the $\ell_1$ norm weighted by the inverse median absolute deviation, such that $MAD_k$ of a feature $k$ over the set of points $P$ is as shown in equation~\ref{eq:mad}.
\begin{equation}
    {MAD}_k = median_{j\in{P}}(|X_{j,k} - median_{l\in{P}}(X_{l,k})|)
    \label{eq:mad}
\end{equation}

This results in a distance $d$ between a synthetic data point $x'$ and original data point $x$ as described by equation 2.
\begin{equation}
    d(x, x') = \sum_{k\in{F}}\frac{|x_k-x'_k|}{MAD_k}
\end{equation}

This distance metric encourages changing only the features that vary in the training set, and discourages changing features with low variance. The architecture for these three methods is illustrated on the left of Figure~\ref{fig:arch}.

We compare a final feature perturbation method from \citet{looveren2019interpretable} which we label \textit{Prototype}. The authors include five different loss terms in their objective to achieve desirable properties of their counterfactuals. $$ L = cL_{pred} + \beta L_1 + L_2 + L_{AE} + L_{proto}$$ $L_{pred}$ is designed to encourage counterfactuals of a different class. $L_1$ and $L_2$ are combined to form an elastic-net regularizer on the feature perturbations for sparse changes. They include $L_{AE}$ as used by \citet{dhurandhar2018tip} to ensure in-sample reconstructions of counterfactuals. Finally, they guide the search for counterfactuals by introducing \textit{prototypes}, which are the average euclidean representation of a class in the latent space defined by the $K$ closest encoded samples to $E(x_0)$. Specifically, for a class $i$ the prototype is defined as $$proto_i = \frac{1}{K}\sum_{k=1}^K E(x_k^i)$$ where $\left\Vert E(x_k^i) - E(x_0)\right\Vert_2 \leq \left\Vert E(x_{k+1}^i) - E(x_0)\right\Vert_2$. $L_{proto}$, defined as the distance to the closest prototype, effectively tries to speed up the search for counterfactuals by pushing $x_0 +\delta$ to the closest prototype. 

\subsection{Experiments}

We analyze each counterfactual generation method on image and tabular data. Specifically, we use the MNIST digit classification dataset \cite{lecun2010mnist} and the Lending Club Loan dataset \cite{lendingclub}. We design separate binary predictions tasks for each dataset. We train classifiers to distinguish between fours vs nines for the MNIST dataset and generate counterfactuals for the opposite class. The classifiers for the loan data predict whether customers will charge-off or fully pay back their loan. We produce counterfactuals for 1,000 customers that are predicted to charge-off (i.e., $p>0.5$). We conduct a visual comparison of the counterfactuals generated by each model type, and use these, along with our evaluation metrics, to draw conclusions about the quality of each method.

\begin{figure}[hbt!]
  \includegraphics[width=0.48\textwidth]{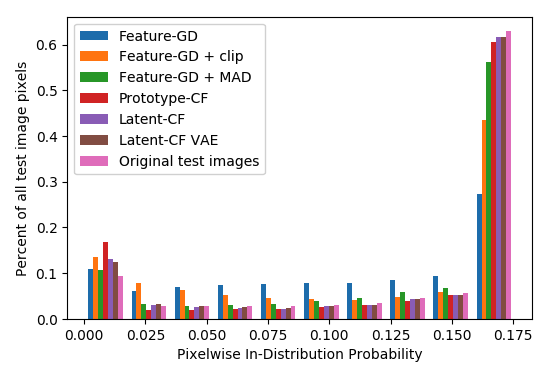}
  \caption{Histogram of Kernel Density Estimate probabilities per pixel computed on original images and counterfactuals. In this case the counterfactual images generated by baseline methods have more low probability pixels than the original images, signifying potential out of distribution changes.}
  \label{fig:kde}
\end{figure}

\subsection{Evaluation Metrics}
In order to evaluate the quality of the counterfactual,  we use three evaluation metrics:\\

\begin{description}[style=nextline]
   \item[In-distribution] The counterfactual should not change the original observation such that proposed features will have a low probability of occurring. We used per-feature kernel density estimation (KDE) to measure the extent to which our counterfactuals are in-sample. Specifically, we estimate the density over intensity values for each feature across the target class population. We compute the probability for each feature given the feature specific KDE and take the average over all features.  Though this does not estimate the full probability of the observation $p(x)$, it serves as a measure of how close each feature is to its own data manifold. For tabular data, we exclude categorical variables from the metric since there is no scenario where the counterfactual can stray from the data manifold of those features. 

An example of this can be seen in Figure~\ref{fig:kde} comparing the probabilities of pixels in counterfactuals of the MNIST dataset to the probabilities of the original pixels.

   \item[Sparsity] The counterfactual should be sparse in the number of changes it makes in the features. We compute the fraction of features that are changed in the generation of the counterfactual. For tabular data we consider the relative magnitudes of change as a fourth metric since sparsity is less informative in low dimensional cases. 
   
   \item[Computational Efficiency] Generating the counterfactual must be computationally efficient. We measure the latency of each method.
\end{description}

\subsection{Models}\label{sec:models}
We use the same pre-trained models across all of our experiments. We train convolutional classifiers, autoencoders, and variational autoencoders for MNIST and train the same models using a dense network for the Lending Club dataset. We selected the top 6 features for the Lending Club data and achieved an AUC of 0.959.

\section{Results and Discussion}

In Table~\ref{tab:mnist_table} and Table~\ref{tab:loan_table}, we present a comparison of each method across all of the proposed metrics. Each of these metrics were calculated using a fixed probability, $p = 0.5$,   as the decision boundary. 

\begin{table}[h!]
\centering
\scalebox{0.94}{
    \begin{tabular}{@{} l *4c @{}}\toprule
    Method        & In-Distribution  & Sparsity (\%)  & Time (s) \\ \midrule
    FGD           & 0.1048         &  96.9       &  \textbf{0.317}        \\ 
    FGD+C         & 0.1121         &  60.0       &     0.394         \\ 
    FGD+MAD       & 0.1314         &  46.5       &     0.407        \\
    Prototype     & 0.1294       & \textbf{15.5} &      7.3         \\ 
    Latent-CF VAE & \textbf{0.1332} &  29.4      &     1.298             \\ 
    Latent-CF     & 0.1326          &  29.7      &     1.380 \\\bottomrule
    \end{tabular}
}
\caption{MNIST counterfactual quality metrics given a target counterfactual class probability of 0.5. }
\label{tab:mnist_table}
\end{table}

\begin{table}[h!]
\centering
\scalebox{0.94}{
    \begin{tabular}{@{} l *4c @{}}\toprule
    Method        & In-Distribution  & Sparsity (\%)  & Time (s) \\ \midrule
    FGD                   & 0.0387         &  83.3     &      0.72        \\ 
    FGD+C            & 0.0387           &  83.3      &      0.74         \\ 
    FGD+MAD   & 0.0387        &  83.3      &      1.09        \\ 
    Prototype                & 0.0407    &  \textbf{74.0} &        2.57         \\ 
    Latent-CF VAE          & 0.0405 &   83.3        &             0.250  
    \\ 
    Latent-CF              & \textbf{0.0431} &      83.3      & \textbf{0.21}  \\\bottomrule     
    
    \end{tabular}
}
\caption{Lending Club counterfactual quality metrics given a target counterfactual class probability of 0.5.}
\label{tab:loan_table}
\end{table}

\begin{figure*}
\scalebox{0.83}{
\begin{tabular}{cccc}
    & Latency & Sparsity & In-Distribution Probability\\
\rotatebox[origin=c]{90}{MNIST} &
\includegraphics[width=6.5cm,valign=c]{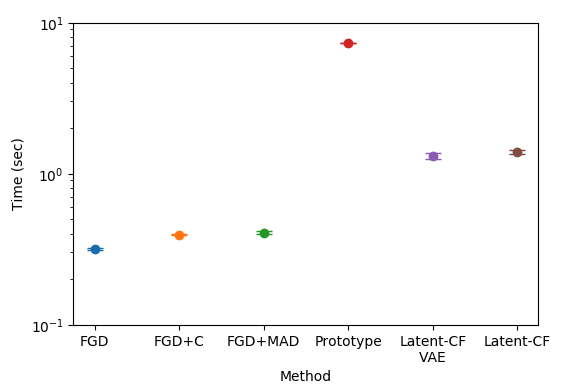}&
\includegraphics[width=6.5cm,valign=c]{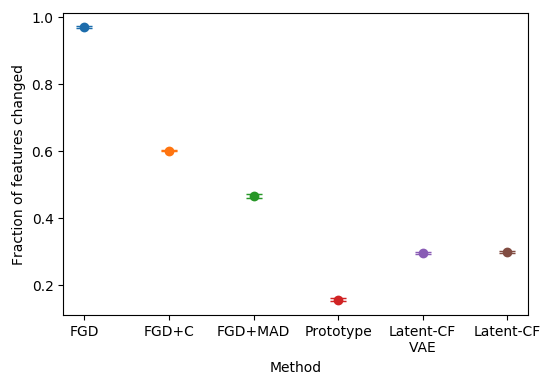}&
\includegraphics[width=6.5cm,valign=c]{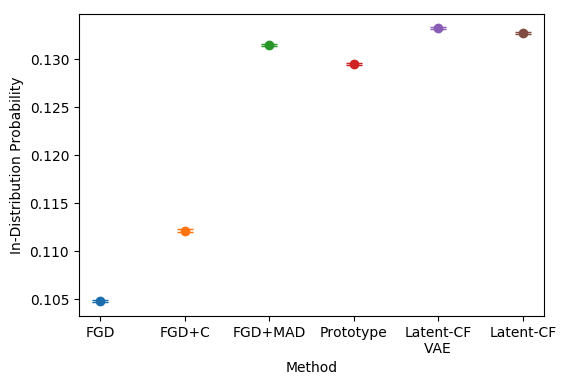}\\

\rotatebox[origin=c]{90}{Lending Club} &
\includegraphics[width=6.5cm,valign=c]{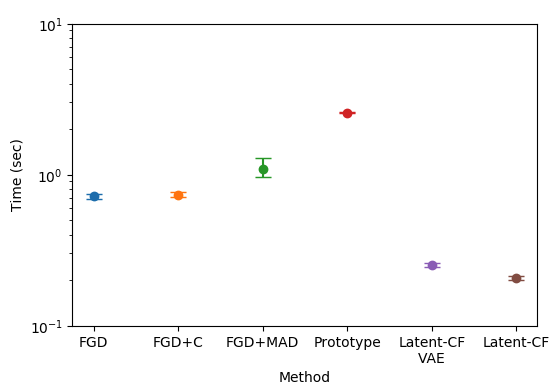}&
\includegraphics[width=6.5cm,valign=c]{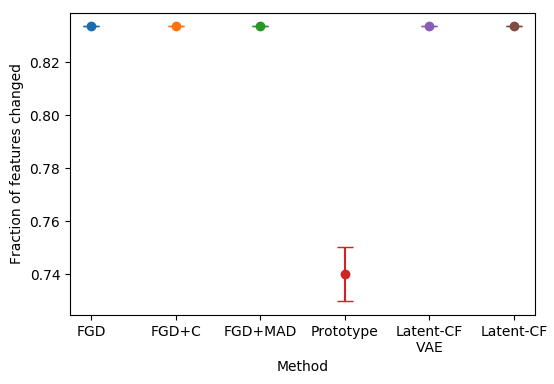}&
\includegraphics[width=6.5cm,valign=c]{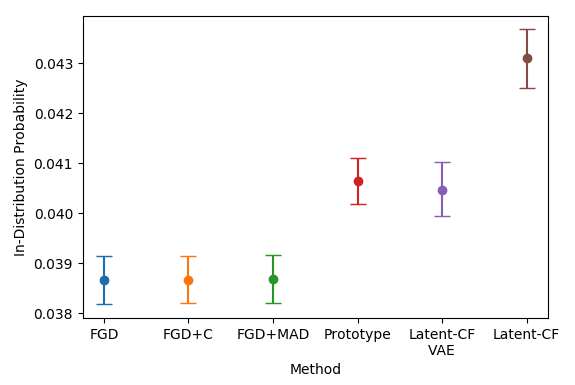}
\end{tabular}
}
\caption{Comparison of counterfactual metrics for MNIST (top) and Lending Club (bottom) for all methods (from left to right in each plot: FGD, FGD+C, FGD+MAD, Prototype, Latent-CF VAE, Latent-CF). Latent-CF methods consistently outperform other methods in generating in-distribution counterfactuals and provide a balance between latency and sparsity (lower is better) of solutions when compared to simple and complex feature perturbation methods. }
\label{fig:exp_graphs}
\end{figure*}
 
\begin{figure*}[hbt!]
    \centering
    \begin{subfigure}{0.95\textwidth}
        \centering
        \includegraphics[width=.98\linewidth]{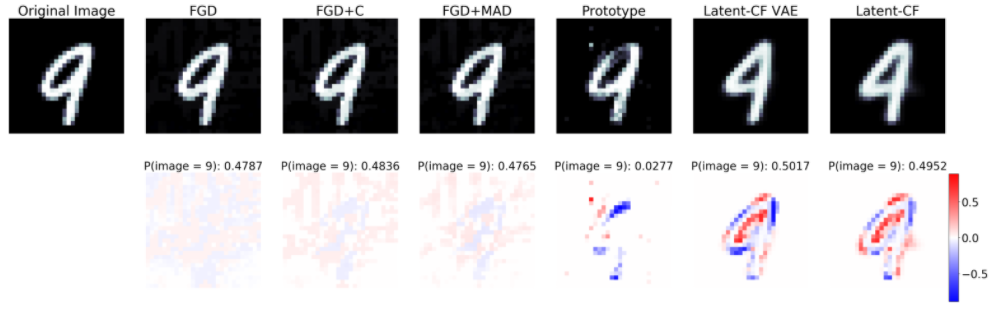}
        \label{fig:cf1}
    \end{subfigure}
    \begin{subfigure}{0.95\textwidth}
        \centering
        \includegraphics[width=.98\linewidth]{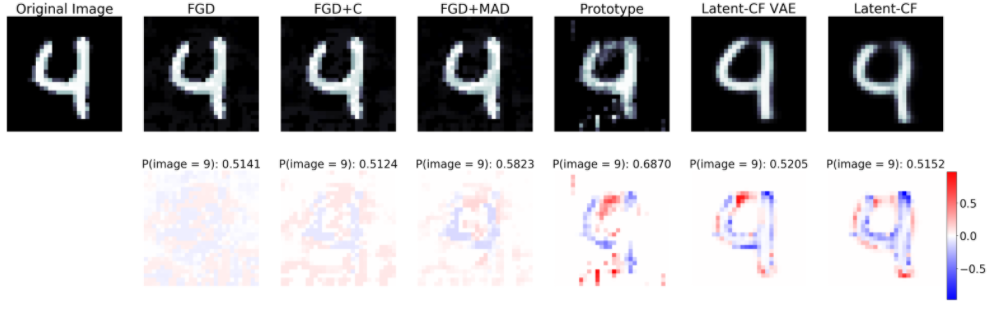}
        \label{fig:cf2}
    \end{subfigure}
    \caption{Original MNIST digits in and corresponding counterfactuals. Heat maps corresponds to the changes in intensities of each pixel. FGD methods produce a multitude of small changes to a a majority of pixels, while Latent-CF only changes pixels around the digits. Prototypes makes limited changes to the digit as well, but tends to make other unnecessary changes near the borders. Prototypes design also allows the counterfactual to overshoot the decision boundary resulting in higher probabilities for the opposite class.    }
    \label{fig:images}
\end{figure*}

 The same results are illustrated in Figure~\ref{fig:exp_graphs} along with 95\% confidence intervals. There are clear trade-offs among the counterfactual methods and differences that arise in results between datasets that we believe are mostly due to disparity in dimensionality. Latency is a prime example of this phenomenon. As we expected, the simplest methods (FGD, FGD+C) are the fastest when it comes to generating image counterfactuals. There are no extra layers to pass gradients through or extra orthogonal objectives slowing down convergence. It also seems much easier to generate adversarial examples to fool the image classifier. This is apparent in Figure~\ref{fig:images} where the heatmaps of FGD and FGD+C are filled with small changes in a majority of pixels. We see the latency advantage disappear with tabular data since it is harder to game the classifier with only six features. In both cases, both Latent-CF methods are almost an order of magnitude faster than Prototypes. We do need to point out that the Latent-CF methods have the advantage of direct automatic differentiation, so we would expect a faster computation time. 

\begin{table}[h!]
\scalebox{0.9}{
\centering
   \begin{tabular}{@{} l *4c @{}}\toprule
    \textit{Loan 1}   & Original &	FGD	 & Prototype & 	Latent-CF \\ \midrule
    Default Prob&	94.6\%	&50.0\%	 & 35.4\%	& 48.5\% \\ \midrule
    dti &	18.8 &	\textcolor{red}{12.4}	& \textcolor{blue}{18.9	} & \textcolor{red}{12.7} \\
    loan\_amnt (\$K)&	22.0 &	\textcolor{red}{13.3} &	22.0 &	\textcolor{red}{12.6} \\
    int\_rate &	9.2 &	\textcolor{red}{7.8} &	\textcolor{blue}{10.3 } &	\textcolor{blue}{11.3} \\
    annual\_inc (\$K)	& 51.0 &	\textcolor{blue}{79.5} & 		\textcolor{blue}{85.6} &		\textcolor{blue}{52.3} \\
    fico	& 589 &	\textcolor{blue}{654}	 & \textcolor{blue}{651}	& \textcolor{blue}{656} \\
    term (months)	& 60 	& 60	& \textcolor{red}{36} &	60 \\
     \bottomrule
    \end{tabular}
}

\bigskip
\scalebox{.9}{
\centering
   \begin{tabular}{@{} l *4c @{}}\toprule
    \textit{Loan 2 }& Original &	FGD	 & Prototype & 	Latent-CF \\ \midrule
    Default Prob&	98.3\%&	49.3\%	&48.9\%	&47.3\% \\ \midrule
    dti	&24.9	&\textcolor{red}{12.5}&	\textcolor{red}{22.3}	&\textcolor{red}{11.8} \\
    loan\_amnt (\$K) &	30.0 &	\textcolor{red}{13.2} &	\textcolor{red}{17.5} &	\textcolor{red}{22.0} \\
    int\_rate &	18.0 &	\textcolor{red}{13.8} &	18.0 &	\textcolor{red}{15.5}  \\
    annual\_inc (\$K) &	88.8 &	\textcolor{blue}{177.8} &	\textcolor{blue}{126.7} &	\textcolor{blue}{99.0} \\
    fico &	544	 & \textcolor{blue}{639} &	\textcolor{blue}{649}  &	\textcolor{blue}{653} \\
    term (months) &	60  &60 &	60 &	60 \\
     \bottomrule
    \end{tabular}
}
\caption{We include two separate loan counterfactual examples from the test set. The original loan application and its associated probability of default (according to the classifier) in the first column. The remaining three columns consist of counterfactual methods' changes to the original loan application to move the probability across the decision boundary. }
\label{tab:loan_examples}

\end{table}

\begin{table*}[h!]
\centering
   \begin{tabular}{@{} l *5c @{}}\toprule
    Method  &  dti &  loan\_amnt & int\_rate &   annual\_inc &  fico\\ \midrule
     FGD     &  -32 (32) &        -37 (37) &       -15 (18) &          58  (59) &                    11 (11)\\
    FGD+C       &  -31 (32)&        -37 (37) &       -15 (18)&          58 (58)&                    11 (11)\\
    FGD+MAD        &  -32 (32)&        -37 (37) &       -15 (18) &          59 (59)&                    11 (11)\\
    Prototype     &    0 (16) &         -9 (25)&        10 (16)&          39 (43) &                    14 (14)\\
    Latent-CF VAE &   16 (23)&         -1 (30)&         2 (22)&          -2 (25)&                    14 (14) \\
    Latent-CF     &  -26 (29)&          4 (26)&        -6 (26)&          24 (32)&                    13 (13) \\ \bottomrule
    \end{tabular}
      
\caption{Average percent change (average absolute percent change) from original loan to counterfactual loan. FGD, FGD+C, FGD+MAD produce more extreme changes and are strongly biased to the direction of change. We see this less with other methods that have some understanding of the data distribution. \textit{fico} is the most salient variable and is consistently increased under all methods.  }
\label{tab:perc_change}
\end{table*}

\begin{figure*}[h!]
\centering
     \includegraphics[width=.99\linewidth]{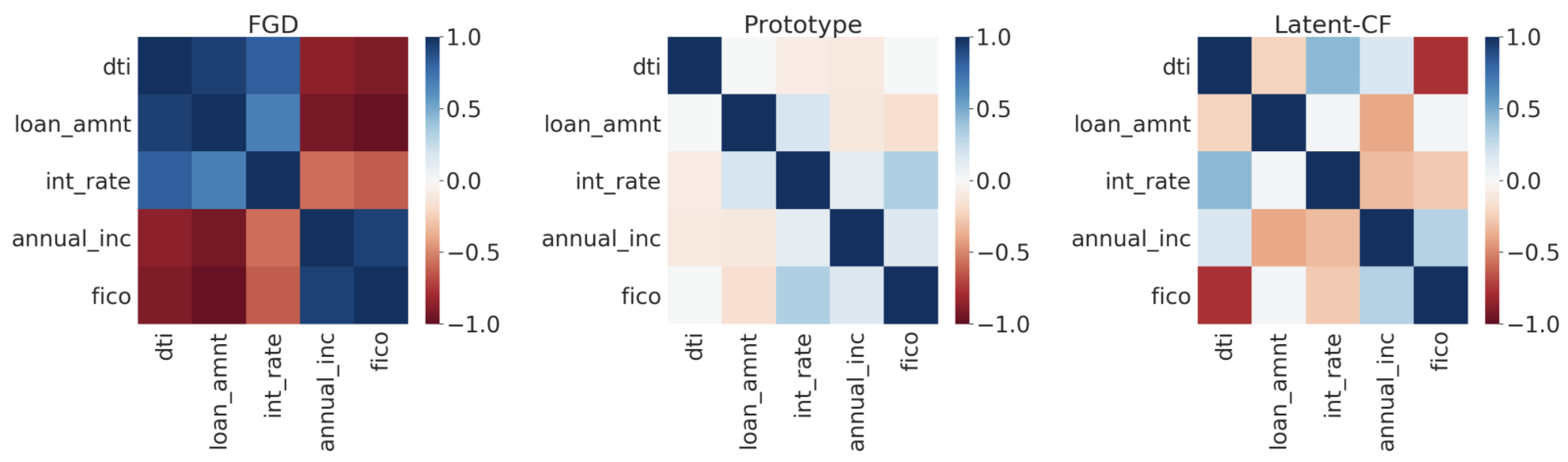}
    \caption{Correlation of feature perturbations in counterfactuals on the Lending Club dataset. FGD jointly decreases the factors that contribute to default and increases the factors that lead to fully paid loans. Prototype and Latent-CF take into account the data distribution and provide varied counterfactuals that are conditioned on the given sample.  }
    \label{fig:loan_corr}
\end{figure*}

The complexity of Prototypes loss function slows convergence but helps it avoid making changes to the original sample. In both datasets, Prototypes makes fewer perturbations. In the case of the MNIST datasets we see large percentages of pixels being changed by feature gradient descent methods, another side affect of the ability to manipulate classifier predictions with many small changes. Latent-CF methods produce rather sparse counterfactuals since autoencoders are not likely to change the bordering black pixels. Since there are so few variables in the tabular data, all the methods except Prototypes change the five continuous variables 100\% of the time and never flip the binary variable (term of the loan -- 36 or 60). Because of the overwhelming effect of FICO score on default probability and the separation between groups with different loan terms in the latent space of autoencoders, gradient descent never pushes these counterfactuals to change the one categorical variable. 

Latent-CF maintains a balance of computational efficiency and sparsity compared to the other baselines and consistently outperforms when it comes to in-distribution counterfactuals in high or low dimensional data. We also note that regularizing the latent space doesn't result in significant improvements in our key metrics, but may be slightly beneficial with high dimensionality. The benefits of Latent-CF are clearly illustrated in the MNIST counterfactual examples in Figure~\ref{fig:images}. The feature gradient descent based methods have small out of distribution changes on the edges of the image data. Prototypes also tends to introduce extraneous pixels surrounding the digits and can overshoot the 0.5 decision boundary, while Latent-CF produces clean, sensible images. 

Finally, we see a clear difference between the behavior of counterfactuals generated for tabular data from simple feature based methods and those produced by Prototypes and Latent-CF. FGD, FGD+C, and FGD+MAD tend to cause larger magnitude changes to the original sample as detailed in Table~\ref{tab:perc_change}. They behave similarly and in a consistent manner: decreasing factors that are associated with loan default and increasing factors associated with fully paid loans. Furthermore, their changes are highly correlated as illustrated in Figure~\ref{fig:loan_corr}. When \textit{dti} is decreased, so are \textit{loan\_amt} and \textit{int\_rate}. Simultaneously, we usually see increases in \textit{annual\_inc} and \textit{fico}. While these changes may align with intuition in general, on a case by case basis it may be more realistic to hold variables constant and make larger changes to others. For example, keep the loan amount stable and decrease the interest rate. Prototypes and Latent-CF (and Latent-CF VAE) are grounded in real examples or the joint distribution of the training data, and thus, are able to act differently depending on the given sample.

\section{Conclusions}
We demonstrate the benefits of performing perturbations in a representative latent space compared to various methods in feature space for counterfactual generation. We show that these methods benefit from sparsity and in-sample perturbations lacking in simpler methods and incur a significant speed up over more complex feature based techniques like Prototypes. The use of variational autoencoders for latent counterfactual generation exhibits little to no benefit over vanilla autoencoders for the current datasets (MNIST and Lending Club).

\section{Future Work}
While Latent-CF provides a good baseline for tabular data using the Lending Club dataset, our tests on higher dimensional data is limited to MNIST. It would be helpful to examine a tabular dataset using more dimensions to decouple the effect of structure and dimensionality on metrics like sparsity and in-distribution. Additionally, we recognize the computational efficiency advantage automatic differentiation gives Latent-CF over model agnostic methods like Prototypes. We hope to examine the extent of performance degradation from optimization methods in latent space without automatic differentiation such as genetic or Bayesian driven techniques.

\bibliography{latent_cf}
\end{document}